\documentclass[runningheads]{llncs}
\usepackage{eccv}

\usepackage{eccvabbrv}
\usepackage{wrapfig}

\usepackage{graphicx}
\usepackage{booktabs}
\usepackage{placeins}
\usepackage{adjustbox}        % 提供 adjustbox 环境
\usepackage{makecell}         % 你用了 \makecell
\usepackage{multirow}         % 你用了 \multirow（截图里有）
\usepackage[table]{xcolor}
\newcommand{\gradSLERIR}{%
\textcolor{blue!90!cyan}{S}%
\textcolor{blue!75!violet}{L}%
\textcolor{violet!45!blue}{E}%
\textcolor{violet!65!magenta}{R}%
\textcolor{magenta!60!violet}{-}%
\textcolor{magenta!70!violet}{I}%
\textcolor{magenta!80!violet}{R}%
}
\usepackage{siunitx}
\usepackage[accsupp]{axessibility}  % Improves PDF readability for those with disabilities.

\usepackage{hyperref}

\usepackage{orcidlink}

\begin{document}

\title{\textbf{\texorpdfstring{\gradSLERIR}{SLER-IR}}: Spherical Layer-wise Expert Routing for All-in-One Image Restoration}


\authorrunning{Peng et al.}
\titlerunning{SLER-IR}

\author{
Shurui Peng$^{1*}$,~~~
Xin Lin$^{2*}$,~~~
Shi Luo$^{1*}$,~~~
Jincen Ou$^{1}$,~~~ \\
Dizhe Zhang$^{3}$,~~~
Lu Qi$^{3}$,~~~
Truong Nguyen$^{2}$,~~~
Chao Ren$^{1\dagger}$~~~
\\[0.2cm]
$^1$Sichuan University~~
$^2$University of California San Diego~~
$^3$Insta360 Research
\\[0.1cm]
$^{*}$Equal contribution. \quad
$^{\dagger}$Corresponding author.
}

\institute{}% TODO FINAL: Replace with your institution list.

\maketitle

\begin{abstract}

Image restoration under diverse degradations remains challenging for unified all-in-one frameworks due to feature interference and insufficient expert specialization. We propose \textbf{\texorpdfstring{\gradSLERIR}{SLER-IR}}, a spherical layer-wise expert routing framework that dynamically activates specialized experts across network layers. To ensure reliable routing, we introduce a Spherical Uniform Degradation Embedding with contrastive learning, which maps degradation representations onto a hypersphere to eliminate geometry bias in linear embedding spaces. In addition, a Global–Local Granularity Fusion (GLGF) module integrates global semantics and local degradation cues to address spatially non-uniform degradations and the train–test granularity gap. Experiments on three-task and five-task benchmarks demonstrate that \textbf{\texorpdfstring{\gradSLERIR}{SLER-IR}} achieves consistent improvements over state-of-the-art methods in both PSNR and SSIM. Code and models will be publicly released.
\end{abstract}

\section{Introduction}
\label{sec:intro}

Image restoration aims to reconstruct high-quality images from degraded inputs (e.g., noise, rain, blur), serving as a fundamental prerequisite for robust downstream vision tasks.
Traditionally, it has been treated as a collection of task-specific problems, with models tailored exclusively for noise, haze, rain, or blur \cite{zhang2017beyond,zhang2018ffdnet,huang2022casapunet,huang2021neighbor2neighbor,zhang2017learning,lin2023multi,cai2016dehazenet,song2023vision,liu2019griddehazenet,chen2020pmhld,yang2025demnet,ren2019progressive,yang2017deep,wang2019erl,jiang2020multi,li2019heavy,chen2021all,qin2024restore,liu2018desnownet,lin2024dual,cho2021rethinking,park2020multi,lin2025re,wei2018deep,guo2020zero,moran2020deeplpf,wu2022uretinex,cai2023retinexformer,dong2015image,lin2023unsupervised}. While effective within their respective domains, these methods lack the flexibility to handle complex or unseen degradations. Extending them to new scenarios requires labor-intensive retraining, limiting their practicality in real-world applications.

\iffalse
\begin{figure}[!t]
  \centering
  \includegraphics[width=\linewidth,keepaspectratio]{Images/1.png}
  \caption{Quantitative comparisons (PSNR\(^{1-5}\)/SSIM\(^{1-5}\)) of four models (AirNet \cite{li2022all}, Restormer \cite{zamir2022restormer}, MoCE-IR \cite{zamfir2025complexity}, and \textbf{\texorpdfstring{\gradSLERIR}{SLER-IR}} (ours)) across five tasks (Derain, Dehaze, Deblur, Low-light, Denoise).}
  \label{fig:teaser}
\end{figure}

\begin{figure}[!t]
  \centering
  \includegraphics[width=\linewidth,keepaspectratio]{Images/ou_7_2.png}
  \caption{Illustration of representative all-in-one image restoration frameworks. (a) Prompt-based methods \cite{li2022all,potlapalli2023promptir,conde2024instructir,luo2023controlling,tian2024instruct} adapts a unified parameter space by modulating features with prompts or contrastive learning. (b) Expert-based methods \cite{luo2023wm,ai2024lora,zamfir2024efficient,yang2024all,yu2024multi,li2020all,zamfir2025complexity} designs to activate task-specific experts. (c) Our Degradation-Perceptive Experts Restormer equipped with a degradation-aware adaptive restoration routing mechanism and a global–local degradation fusion module, enabling robust and specialized restoration.}
  \label{fig:compare1}
\end{figure}
\fi

\begin{figure*}[t]
  \centering

  % Left (原 Right)
  \begin{minipage}[t]{0.48\linewidth}
    \vspace{0pt}
    \centering
    \includegraphics[width=\linewidth]{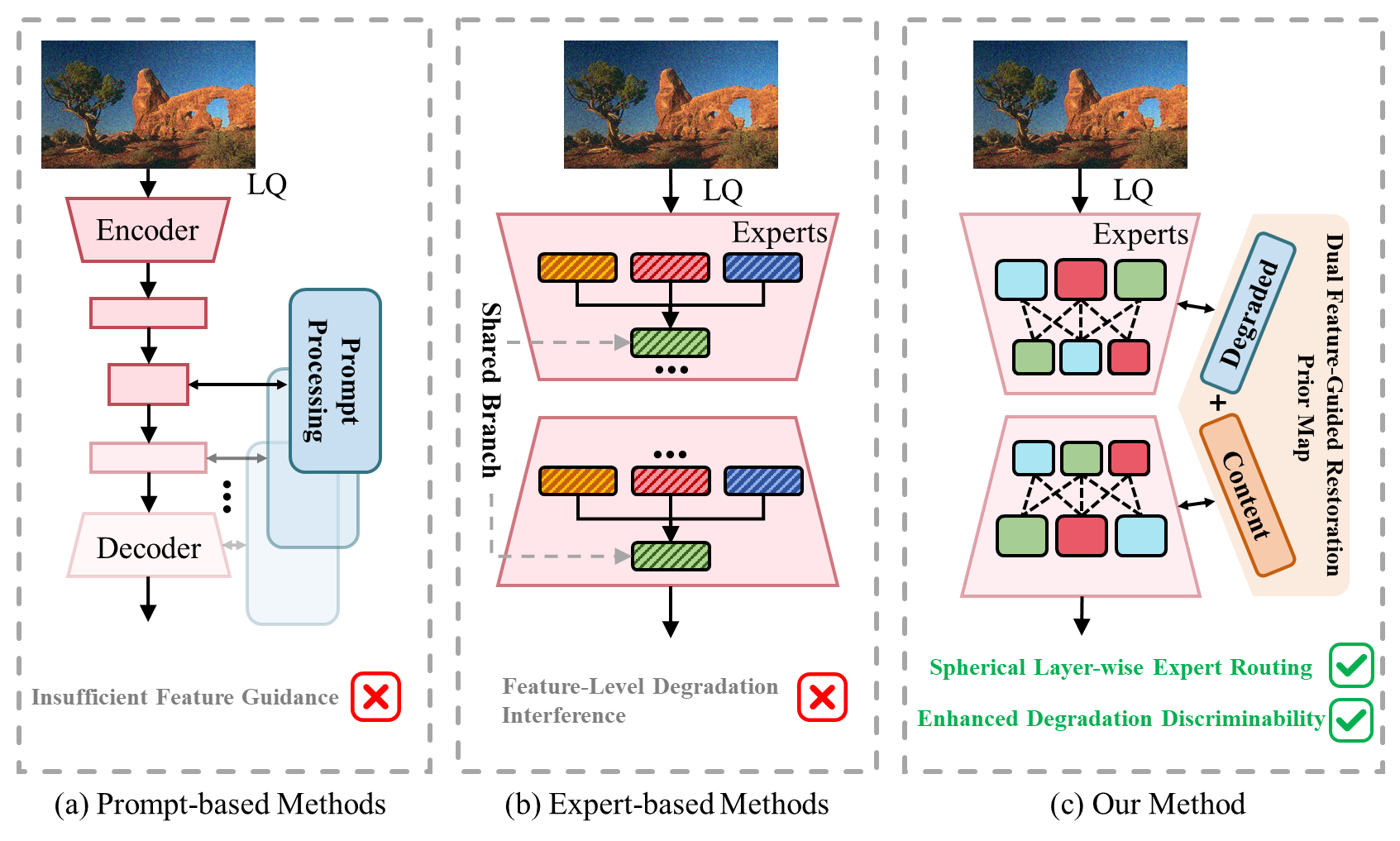}
    \caption{Illustration of representative all-in-one image restoration frameworks. (a) Prompt-based methods~\cite{li2022all,potlapalli2023promptir,conde2024instructir,luo2023controlling,tian2024instruct} adapt a unified parameter space via feature modulation. (b) Expert-based methods~\cite{luo2023wm,ai2024lora,zamfir2024efficient,yang2024all,yu2024multi,li2020all,zamfir2025complexity} activate task-specific experts. (c) Our \textbf{\texorpdfstring{\gradSLERIR}{SLER-IR}} introduces spherical expert routing and global–local granularity fusion for robust restoration.}
    \label{fig:compare1}
  \end{minipage}
  \hfill
  % Right (原 Left)
  \begin{minipage}[t]{0.48\linewidth}
    \vspace{-2mm}
    \centering
    \includegraphics[width=\linewidth]{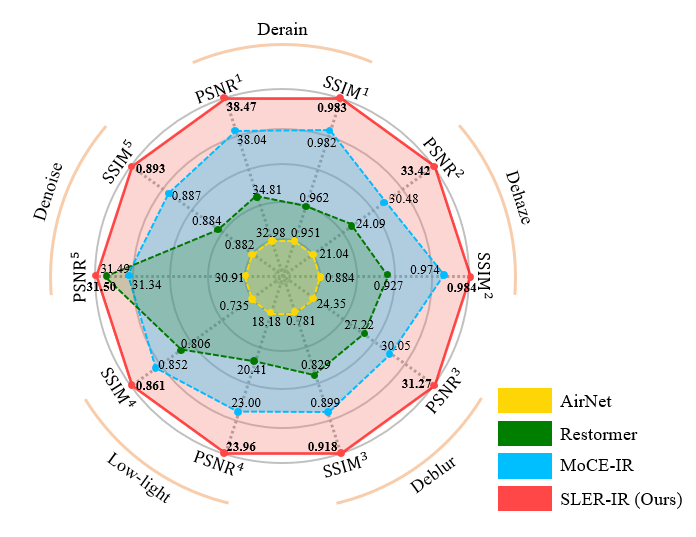}
    \caption{Quantitative comparisons (PSNR$^{1\text{-}5}$/SSIM$^{1\text{-}5}$) of four models (AirNet~\cite{li2022all}, Restormer~\cite{zamir2022restormer}, MoCE-IR~\cite{zamfir2025complexity}, and \textbf{\texorpdfstring{\gradSLERIR}{SLER-IR}} (ours)) across five image restoration tasks (Derain, Dehaze, Deblur, Low-light, Denoise).}
    \label{fig:teaser}
  \end{minipage}

\end{figure*}

To address this, all-in-one restoration frameworks \cite{li2022all,potlapalli2023promptir,conde2024instructir,luo2023controlling,zhang2025perceive,li2023prompt,tian2024instruct,zhang2023ingredient,zeng2025vision,cui2023image,valanarasu2022transweather} have emerged to handle diverse degradations within a single model, offering superior generalization across real-world conditions. As illustrated in Figure \ref{fig:compare1}, these methods generally fall into two paradigms: feature modulation (Figure \ref{fig:compare1}(a)), which employs prompts or contrastive learning to adapt a unified parameter space (e.g., AirNet \cite{li2022all}, PromptIR \cite{potlapalli2023promptir}, InstructIR \cite{conde2024instructir}, DA-CLIP \cite{luo2023controlling}), and architectural adaptation (Figure \ref{fig:compare1}(b)), which utilizes Mixture-of-Experts (MoE) to activate task-specific components (e.g., WM-MoE \cite{luo2023wm}, MoCE-IR \cite{zamfir2025complexity}).
However, both paradigms exhibit notable limitations. Firstly, feature modulation methods suffer from feature interference within the shared backbone. Conflicting objectives, exemplified by the trade-off between suppressing noise and restoring high-frequency details, hinder fine-grained, degradation-specific modeling. Secondly, expert-based architectures often restrict experts to local modules, failing to fully exploit degradation priors. Furthermore, existing routing mechanisms struggle to resolve adversarial relationships among tasks, leading to suboptimal specialization. Beyond these structural issues, most methods overlook locally non-continuous degradations (e.g., rain streaks) and neglect the granularity gap between training crops and full-resolution inference.

To address these limitations, we propose \textbf{\texorpdfstring{\gradSLERIR}{SLER-IR}}, an all-in-one image restoration framework equipped with a spherical layer-wise expert routing mechanism. Structurally, we transform the encoder–decoder by replacing each backbone block with multiple parameter-independent experts. 
To ensure precise expert selection, we employ a spherically uniform embedding strategy. This approach mitigates the class-distance bias often found in linear embeddings and, complemented by a triplet-based contrastive loss, significantly enhances degradation discriminability. 
Consequently, the network performs layer-wise dynamic activation, adaptively recalibrating its processing strategy as features evolve from global to local levels.
As a result, \textbf{\texorpdfstring{\gradSLERIR}{SLER-IR}} enables a compositional inference paradigm, where expert specialization and path diversity emerge naturally through layer-wise routing. Such exponentially diverse routing trajectories allow the model to flexibly adapt to complex and composite degradations without increasing inference overhead.
Finally, a Global–Local Granularity Fusion module bridges the granularity gap between training patches and full-resolution inference images, ensuring consistent and fine-grained guidance. 
Extensive experiments show that our proposed method achieves comprehensive superiority across multiple types of degradation, as highlighted in the quantitative comparison in Figure \ref{fig:teaser}.

The main contributions of this work are summarized as follows:

\begin{enumerate}

\item We propose \textbf{\texorpdfstring{\gradSLERIR}{SLER-IR}}, a unified all-in-one image restoration framework featuring Spherical Layer-wise Expert Routing. By expanding each backbone layer into parameter-independent experts and enabling layer-wise expert activation, the model supports compositional inference paths and achieves progressive specialization across diverse and composite degradations.

\item We introduce Spherical Uniform Degradation Embedding with Contrastive Learning for fine-grained degradation discrimination. By mapping degradation embeddings onto a unit hypersphere and optimizing them with a triplet-constrained contrastive objective, our method mitigates class-distance bias in linear spaces, thereby enhancing the separability and robustness of degradation representations for stable routing.

\item We design Global–Local Granularity Fusion (GLGF) to harmonize global image context (CLS tokens) with local degradation cues (patch tokens). GLGF bridges the granularity gap between patch-based training and full-resolution inference, leading to robust routing guidance under complex and spatially non-uniform degradations.

\end{enumerate}

\section{Related Works}
\label{sec:related}

\textbf{Single-Task Image Restoration.}
Single-task image restoration methods target specific degradation types and typically assume that the degradation pattern is known and consistent with the training data. While these approaches achieve strong performance within their respective domains, they often lack the flexibility to generalize to unseen or mixed degradations. Existing methods can be categorized by degradation type. For denoising, representative works include DnCNN \cite{zhang2017beyond}, FFDNet \cite{zhang2018ffdnet}, CasaPuNet \cite{huang2022casapunet}, and Neighbor2neighbor \cite{huang2021neighbor2neighbor}. 
For dehazing, typical methods include DehazeNet \cite{cai2016dehazenet}, FFA-Net \cite{qin2020ffa}, GridDehazeNet \cite{liu2019griddehazenet}, and PMHLD \cite{chen2020pmhld}. 
For deraining, representative models include PReNet \cite{ren2019progressive} and ERL-Net \cite{wang2019erl}. 
DesnowNet \cite{liu2018desnownet} addresses image desnowing, while DeblurGAN \cite{kupyn2018deblurgan} and DeblurGANv2 \cite{kupyn2019deblurgan} focus on deblurring. 
For low-light enhancement, representative works include LIME \cite{guo2016lime}, DeepLPF \cite{moran2020deeplpf}, Uretinex-Net \cite{wu2022uretinex}, and RetinexFormer \cite{cai2023retinexformer}. 
In super-resolution, typical methods include SRCNN \cite{dong2015image} and EDSR \cite{lim2017enhanced}.

More recently, several general-purpose restoration frameworks, such as MPRNet \cite{zamir2021multi}, SwinIR \cite{liang2021swinir}, NAFNet \cite{chen2022simple}, and Restormer \cite{zamir2022restormer}, have been proposed to improve model generality across multiple restoration tasks. Despite their strong performance, these models still require retraining or task-specific adaptation when encountering unseen or composite degradations. To overcome this limitation and better address complex real-world scenarios, all-in-one image restoration frameworks have recently emerged.

\begin{figure*}[!t]
  \centering
  \includegraphics[width=\linewidth,keepaspectratio]{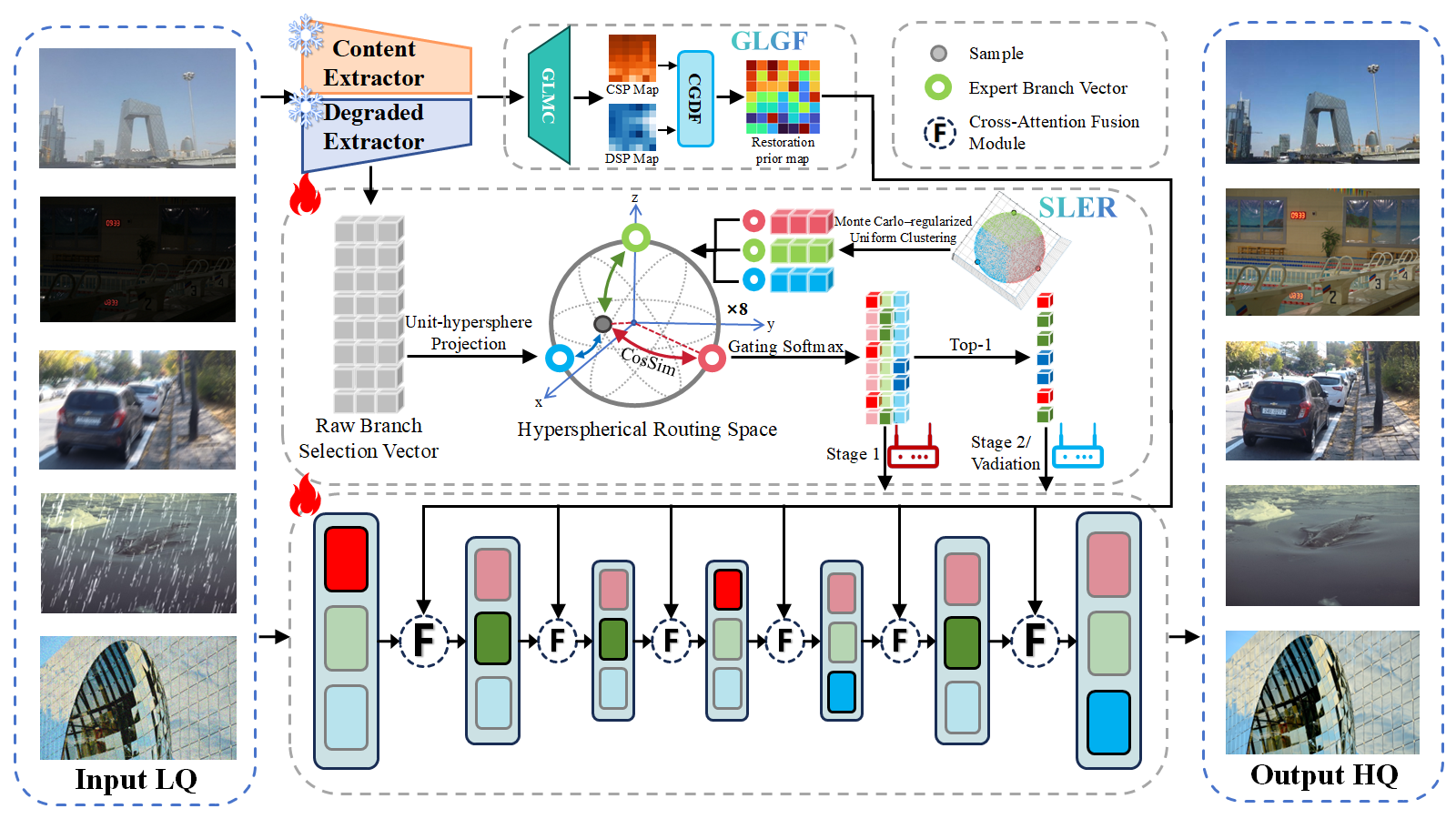}
  \caption{Overview of the proposed \textbf{\texorpdfstring{\gradSLERIR}{SLER-IR}} framework. Given an input LQ image, the degradation extractor produces a raw routing vector, which is projected onto the unit hypersphere for cosine-similarity gating, enabling layer-wise expert selection under Stage~1 (Probabilistic Routing) and Stage~2 (Deterministic Routing). In parallel, a Global–Local Map Construction (GLMC) module derives a content semantic patch (CSP) map and a degradation severity patch (DSP) map, which are fused through Content-Guided Degradation Fusion (CGDF) to guide restoration.
  }
  \label{fig:Framework} 
\end{figure*}

\noindent \textbf{All-in-One Image Restoration.} All-in-one restoration aims to train a unified model capable of handling multiple degradations (e.g., noise, blur, haze, and low-light) without explicit task identification or separate fine-tuning. 

One research direction employs intrinsic or extrinsic degradation prompts to modulate features within a unified parameter space. AirNet \cite{li2022all} learns degradation representations via a contrastive encoder, while PromptIR \cite{potlapalli2023promptir} and InstructIR \cite{conde2024instructir} use prompt-based or language-guided embeddings to steer restoration. DA-CLIP \cite{luo2023controlling} decouples degradation knowledge from content, and UniProcessor \cite{duan2024uniprocessor} further achieves explicit prompt-driven modulation.
Another direction focuses on architectural designs tailored for heterogeneous degradations. Early works, such as \cite{li2020all}, adopt multi-encoder–single-decoder structures, while IPT \cite{chen2021pre} employs a shared backbone with multi-head and multi-tail branches. MoE-based methods—including WM-MoE \cite{luo2023wm}, LoRA-IR \cite{ai2024lora}, DaAIR \cite{zamfir2024efficient}, AMIR \cite{yang2024all}, MEASNet \cite{yu2024multi}, and MoCE-IR \cite{zamfir2025complexity}—activate task-specific experts via routing or gating, with the latter introducing complexity-aware routing.
Despite their progress, these approaches still suffer from limited use of degradation priors and incomplete parameter sharing. Experts often differ only in isolated modules, restricting their adaptability across diverse degradations, and existing balanced-loss strategies inadequately address directional conflicts, hindering effective routing and expert specialization.

\section{Method}
\label{sec:method}

As illustrated in Fig.~\ref{fig:Framework}, \textbf{\texorpdfstring{\gradSLERIR}{SLER-IR}} comprises four integral components: Spherical Layer-wise Expert Routing (SLER) (Sec.~\ref{sec:3.1}), which serves as the core routing backbone and constructs a unified multi-expert architecture with compositional inference paths; Router Design and Similarity-based Expert Gating (Sec.~\ref{sec:3.2}), which performs layer-wise expert selection via similarity-based gating; Hyperspherical Degradation Representation Learning (Sec.~\ref{sec:3.3}), which learns geometry-consistent degradation embeddings to mitigate class-distance bias and stabilize routing decisions; and Global–Local Granularity Fusion (GLGF) (Sec.~\ref{sec:3.4}), which integrates global CLS tokens and local patch cues to bridge the train--test granularity gap and provide robust routing guidance under varying degradations.

\subsection{Spherical Layer-wise Expert Routing}
\label{sec:3.1}

To enable degradation-specific processing within a unified model, we transform the encoder--decoder by expanding each backbone layer into multiple parameter-independent expert nodes. 
Unlike conventional static networks, each layer contains three candidate experts with distinct latent feature representations. 
This multi-expert backbone, governed by Spherical Layer-wise Expert Routing (SLER), forms the core routing architecture of \textbf{\texorpdfstring{\gradSLERIR}{SLER-IR}}.

\begin{wrapfigure}{r}{0.48\linewidth}
  \vspace{-10mm}
  \centering
  \includegraphics[width=\linewidth]{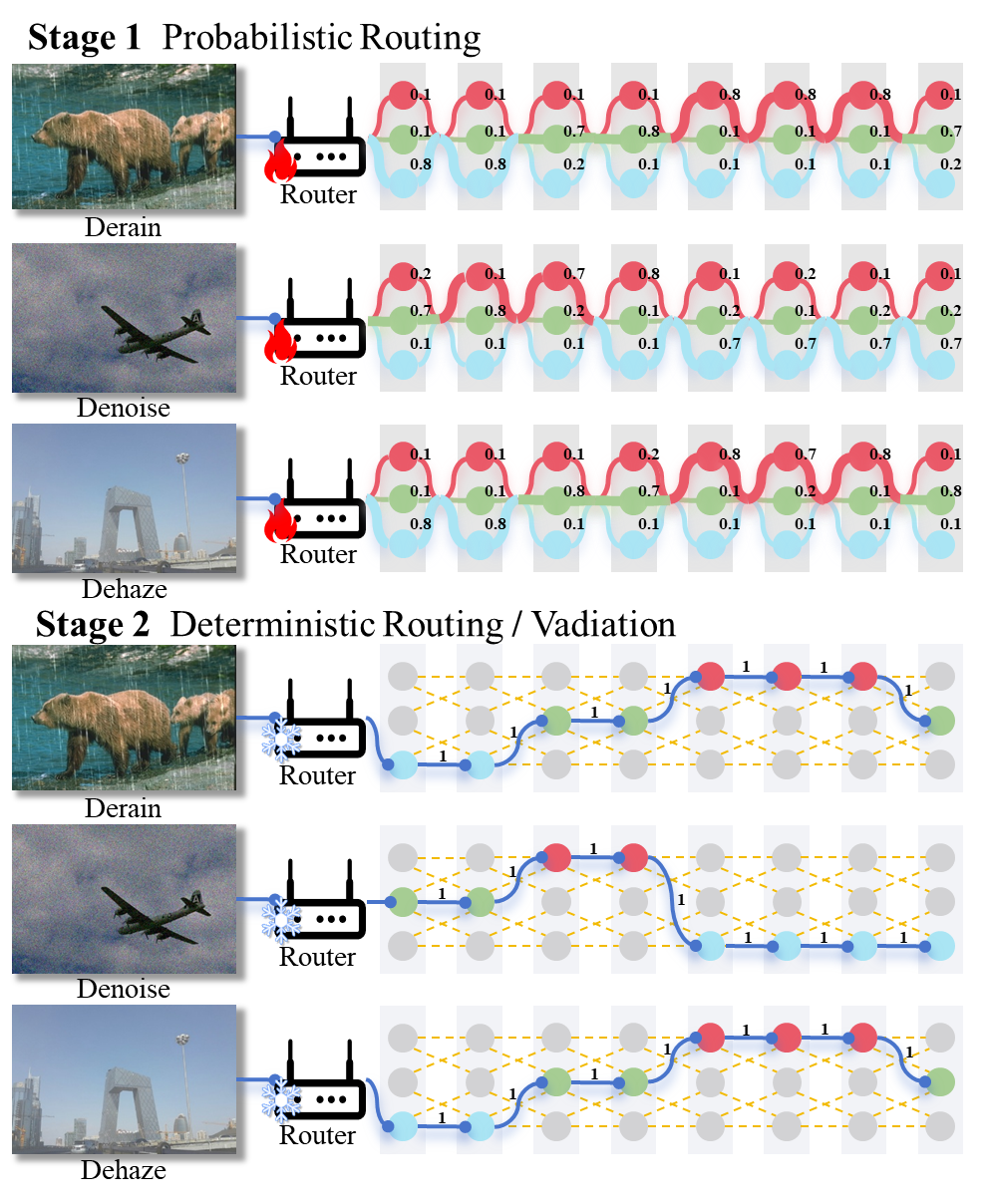}
  \caption{Visualization of dynamic routing trajectories across different degradations. Stage 1 learns degradation-aware routing via probabilistic expert selection, while Stage 2 forms specialized expert paths through deterministic routing across layers for different degradations.}
  \label{fig:route_selection}
  \vspace{-10mm}
\end{wrapfigure}
As shown in Fig.~\ref{fig:route_selection}, during inference the network does not activate all experts; instead, it dynamically selects a specific expert path for each layer based on the input's degradation characteristics. Since routing decisions are made independently across the eight backbone layers, the network can compose up to $3^8 = 6561$ unique inference paths. 
This exponential path diversity allows the model to flexibly specialize to complex and composite degradations without increasing the inference computational burden. 
Importantly, the routing trajectories are not fixed per task but adapt progressively across layers, allowing composite degradations to activate hybrid expert combinations.

While the proposed multi-expert backbone provides exponential path diversity and compositional inference capability, its effectiveness critically depends on how experts are selected. 
Without a reliable degradation-aware routing signal, the large routing space may lead to unstable specialization or suboptimal path selection. 
Therefore, we next introduce the routing mechanism that enables stable and progressive expert activation.

\subsection{Router Design and Similarity-based Expert Gating}
\label{sec:3.2}

\paragraph{Router Inputs and Routing Vector.}
Given an input degraded image $x$, we extract semantic priors using a Content CLIP encoder and degradation cues using a Degraded CLIP encoder. The degradation-related features are then fed into a Router to produce a layer-wise routing representation. Concretely, we take the CLS token from Degraded CLIP and pass it through an MLP followed by L2 normalization to obtain the raw branch selection vector $f \in \mathbb{R}^{K \times d}$:
\begin{equation}
    f = \mathrm{L2}\!\left(\mathrm{MLP}\!\left(\mathrm{CLIP}(x)\right)\right),
\end{equation}
where $K$ is the number of backbone layers (e.g., $K{=}8$) and $d$ is the routing embedding dimension. 
Importantly, the L2 normalization maps routing features onto the unit hypersphere, forming our \emph{spherical uniform embedding} and enabling stable cosine-similarity-based expert selection.

\paragraph{\textbf{Probabilistic Routing} (Stage I: Router Optimization).}
In the first stage, we adopt probabilistic routing to optimize the Router and learn stable degradation-aware representations. 
We compute cosine similarities between each layer-wise routing vector $f_i \in \mathbb{R}^{d}$ and a set of expert centers $\mathbf{C}\in\mathbb{R}^{C\times d}$ (one center per candidate expert), and apply a \emph{row-wise} Softmax to obtain selection probabilities $\mathbf{p}\in\mathbb{R}^{K\times C}$:
\begin{equation}
    \mathrm{Sim}_{i,j}=\frac{f_i\cdot \mathbf{C}_j^{\top}}{\lVert f_i\rVert_2\,\lVert \mathbf{C}_j\rVert_2},\quad
    \mathbf{p}_{i,:}=\mathrm{Softmax}(\mathrm{Sim}_{i,:}),
\end{equation}
where $\mathbf{C}_j\in\mathbb{R}^{d}$ denotes the $j$-th center and $\mathbf{p}_{i,:}\in\mathbb{R}^{C}$ sums to 1. 
The expert centers $\mathbf{C}$ are updated by Monte Carlo--regularized uniform clustering: we perform Monte Carlo sampling over hyperspherical routing embeddings and impose a uniformity constraint, encouraging a more balanced assignment of samples across centers and yielding well-separated, uniformly distributed expert prototypes on the unit hypersphere.
During this stage, the restoration objective (e.g., $L_1$ loss) is jointly optimized with a triplet-constrained contrastive loss to enhance the discriminability of degradation representations on the hypersphere. 
The probabilistic formulation allows all experts to receive gradients, preventing premature specialization and stabilizing routing learning.

\paragraph{\textbf{Deterministic Routing} (Stage II: Restoration Refinement).}
In the second stage, we switch to deterministic routing and freeze the Router to obtain fixed expert assignments. 
Specifically, the expert index with the maximum probability is selected at each layer:
\begin{equation}
    y_i=\arg\max_{j\in\{1,2,\dots,C\}} \mathbf{p}_{i,j},
\end{equation}
where $\mathbf{y}\in\mathbb{N}^{K\times 1}$ denotes the hard selection indices across layers. 
At each backbone layer $l$, only the expert corresponding to $y_l$ is activated, while the Router parameters remain frozen. 
Under fixed routing trajectories, the restoration network is further optimized to refine expert specialization and improve reconstruction quality. Because routing decisions are made independently across depths, the network dynamically recalibrates its processing strategy as feature representations evolve from low-level textures to high-level structures, yielding progressive specialization without activating all experts.

However, the effectiveness of deterministic routing ultimately depends on the quality of the learned degradation representations. 
This raises a fundamental question: what representation structure is required to ensure stable and reliable expert selection? 
We address this in Sec.~\ref{sec:3.3}.

\subsection{Hyperspherical Degradation Representation Learning}
\label{sec:3.3}

\begin{figure*}[t]
  \centering
  \includegraphics[width=\linewidth,keepaspectratio]{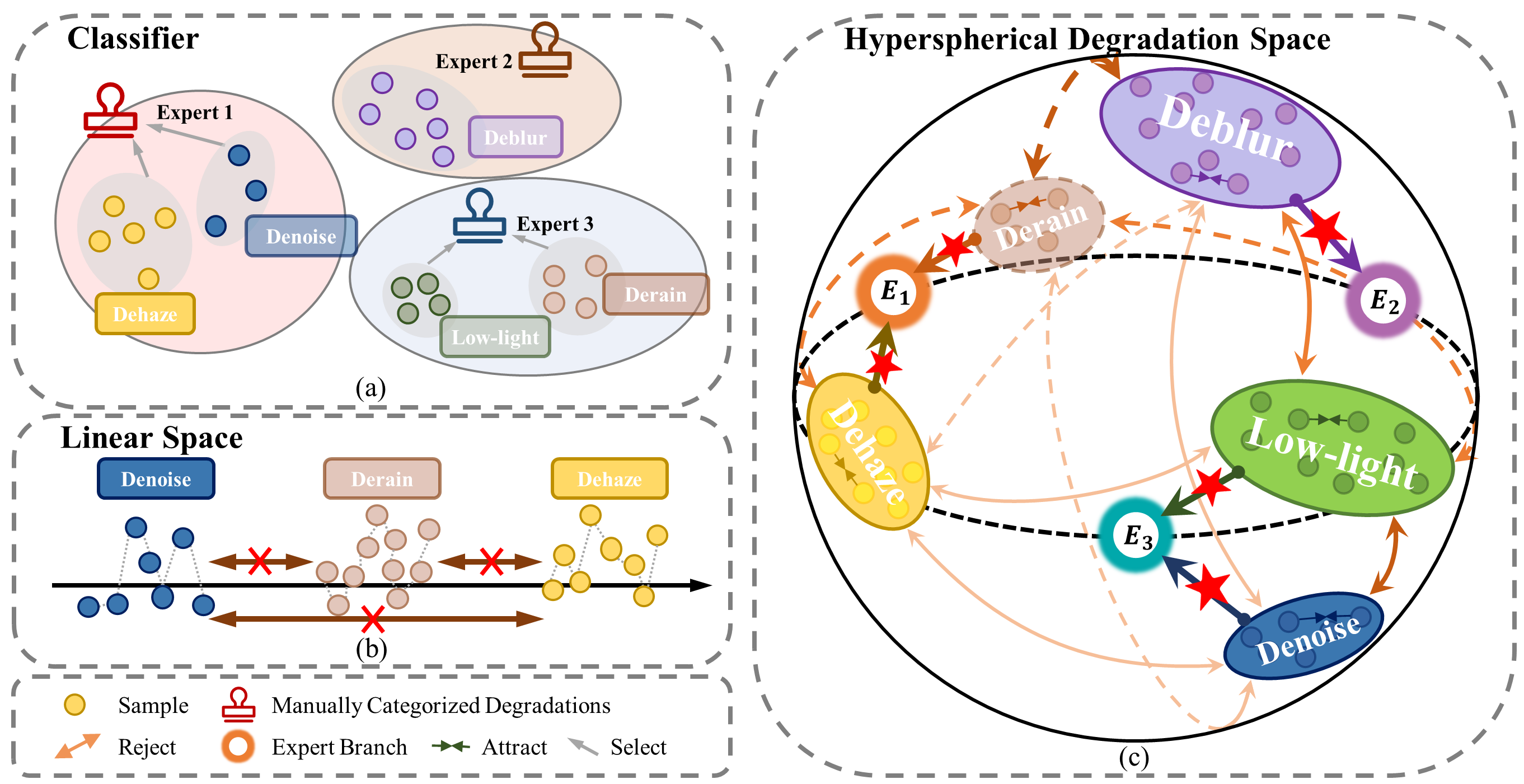}
  \caption{Comparison of existing degradation routing methods and our hyperspherical degradation representation-based branch selection strategy. (a) Traditional expert-based categorization and assignment neglect inter-degradation similarities. (b) Linear-space embeddings may introduce class-distance bias. (c) Our method maps degradations onto a unit hypersphere, enabling geometrically balanced distributions and robust branch selection via cosine similarity.}
  \label{fig:compare2}
\end{figure*}

\paragraph{\textbf{Motivation.}}
Our router performs similarity-based, degradation-aware expert gating (Sec.~\ref{sec:3.2}). 
Hence, routing quality hinges on whether the degradation representation space provides a geometry-consistent similarity structure: expert selection should reflect true degradation affinity rather than artifacts induced by embedding geometry. 
We observe that commonly used linear embedding strategies can introduce biased inter-degradation geometry, which directly destabilizes similarity-based gating.

\paragraph{\textbf{Limitations of Existing Routing Methods.}}
We identify two key limitations:

\noindent \textbf{(i) Rigid mapping and coarse supervision.}
As shown in Fig.~\ref{fig:compare2}(a), many schemes enforce rigid degradation-to-expert assignments (e.g., manually categorized degradations or task-ID gating). When the number of experts is smaller than the number of degradation types, distinct degradations must be merged into the same expert, ignoring fine-grained similarities and differences. This limitation becomes more severe for composite degradations, where the optimal processing path should be progressively composed across layers rather than fixed to a single expert.

\noindent \textbf{(ii) Geometry bias in unconstrained linear embedding spaces.}
As illustrated in Fig.~\ref{fig:compare2}(b), representations learned in an unconstrained linear space often exhibit class-distance bias: inter-class distances/similarities are not explicitly regularized and can become highly uneven. Since our gating is similarity-based, such bias contaminates routing scores and may trigger unstable or incorrect expert selection.

\paragraph{Counterexample aligned with Fig.~\ref{fig:compare2}(b): distance bias misleads similarity gating.}
Consider three degradation types $A$ (Denoise), $B$ (Derain), and $C$ (Dehaze) embedded in a linear space. A biased yet plausible configuration is
\begin{equation}
d(A,C) \gg d(A,B),\ d(B,C),
\label{eq:linear_bias}
\end{equation}
meaning Denoise and Dehaze are artificially pushed far apart while Derain lies in between. Under a similarity-based gating rule
\begin{equation}
j^{*}=\arg\max_{j\in\{1,\dots,C\}} \mathrm{sim}\!\left(\mathbf{f}, \mathbf{c}_j\right),
\label{eq:routing_score_sim}
\end{equation}
the embedding bias in Eq.~\eqref{eq:linear_bias} exaggerates dissimilarity between $A$ and $C$ regardless of their actual restoration affinity. As a result, the router may over-separate degradations that share processing components, or over-merge those that should be separated, leading to branch-selection flips and cross-degradation interference.

\vspace{0.4em}
\noindent \textbf{Spherical Uniform Embedding.}
To eliminate geometry-induced bias, we introduce Spherical Uniform Embedding, mapping degradation representations onto a unit hypersphere and performing angle-based comparison. Given a degradation feature vector $\mathbf{f}$ extracted by the degradation encoder, we project it onto the hypersphere via $L_2$ normalization:
\begin{equation}
    \hat{\mathbf{f}} = \frac{\mathbf{f}}{\|\mathbf{f}\|_2}.
    \label{eq:sue_norm}
\end{equation}
Routing is then performed by cosine similarity to expert centers:
\begin{equation}
    s_j=\cos(\hat{\mathbf{f}},\hat{\mathbf{c}}_j)=\hat{\mathbf{f}}^{\top}\hat{\mathbf{c}}_j,
    \quad j^{*}=\arg\max_j s_j.
\end{equation}
Importantly, we optimize expert centers to be mutually dissimilar (low pairwise cosine similarity), encouraging uniformly distributed expert directions on the hypersphere (spherical-code-like), which increases the minimum inter-center angle and stabilizes expert gating.

\vspace{0.4em}
\noindent \textbf{Hyperspherical Contrastive Optimization.}
To sharpen angular decision boundaries on the hypersphere, we adopt a triplet-constrained contrastive objective. For each layer $l$, similarity between samples $i$ and $j$ is measured by the dot product of normalized vectors (equivalent to cosine similarity on the unit hypersphere):
\begin{equation}
    S_{i,j,l} = \hat{\mathbf{f}}_{i,l}^{\top}\hat{\mathbf{f}}_{j,l}.
    \label{eq:similarity}
\end{equation}
We define the hyperspherical contrastive loss as
\begin{equation}
    L_{\text{HC},l} = \max\!\left( \mathbb{E}_{N}\!\left[S_{i,j,l}\right] - \mathbb{E}_{P}\!\left[S_{i,j,l}\right] + \lambda,\, 0 \right),
    \label{eq:contrastive_loss}
\end{equation}
where $\lambda$ is a margin, and $\mathbb{E}_{P}$/$\mathbb{E}_{N}$ denote the average similarity over positive/negative pairs. The overall objective combines restoration and contrastive terms:
\begin{equation}
    L = L_1 + \alpha \cdot L_{\text{HC}}.
    \label{eq:total_loss}
\end{equation}
This joint optimization yields embeddings that are compact within class and well-separated across classes in angular space, producing more reliable routing scores and reducing cross-degradation interference. To the best of our knowledge, such hyperspherical (spherical-uniform) degradation embedding explicitly designed for expert routing remains underexplored in the image restoration literature, where representations are typically learned in unconstrained linear spaces.

\subsection{Global--Local Granularity Fusion}
\label{sec:3.4}

\begin{wrapfigure}{r}{0.48\linewidth}
  \vspace{-8mm}
  \centering
  \includegraphics[width=\linewidth,keepaspectratio]{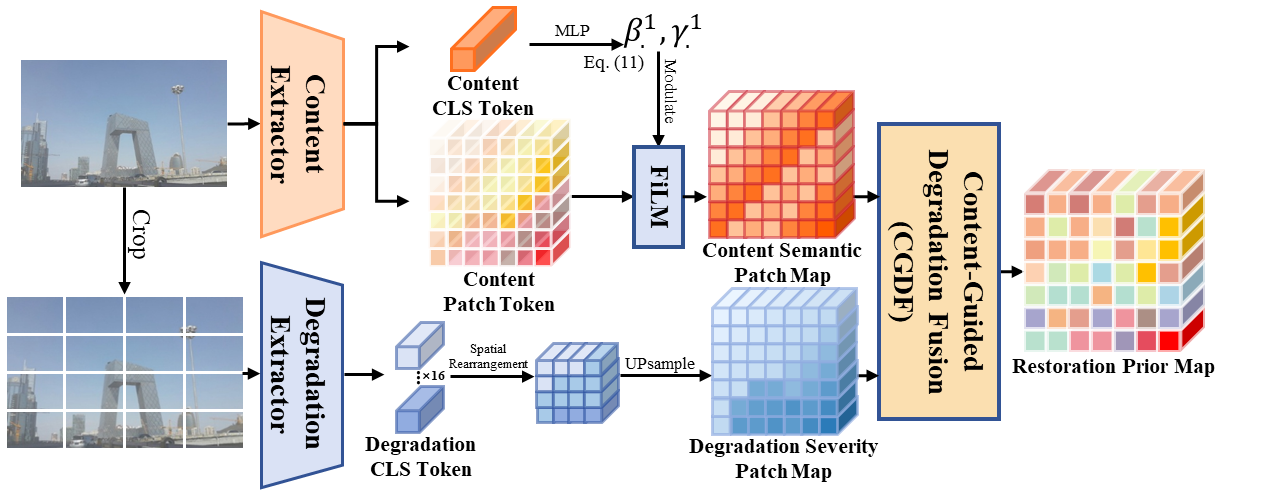}
  \caption{Architecture of GLGF. GLGF consists of GLMC (Global--Local Map Construction) that derives a CSP map (content semantic patch map) and a DSP map (degradation severity patch map), and CGDF (Content-Guided Degradation Fusion) that fuses them into a \emph{restoration prior map} for robust restoration under spatially varying degradations.}
  \label{fig:glfusion}
  \vspace{-10mm}
\end{wrapfigure}

While global representations provide scene-level consistency, restoration under spatially non-uniform degradations requires guidance that is both \emph{content-aware} and \emph{spatially localized}. In real scenarios, corruptions such as rain streaks, localized haze, or mixed degradations often appear only in certain regions. Moreover, patch-based augmentation during training and full-image inference at test time may introduce a granularity mismatch, where local degradation patterns observed in crops are not consistently reflected when processing the entire image.

To address this, we propose Global--Local Granularity Fusion (GLGF), which constructs a \emph{content-guided, degradation-conditioned} restoration prior by aligning global content context with localized degradation evidence. As shown in Fig.~\ref{fig:glfusion}, GLGF includes two components: (i) GLMC to construct two aligned patch-level maps (CSP/DSP), and (ii) CGDF to fuse them into a restoration prior map.

\paragraph{Global--Local Map Construction (GLMC).}
Given an input image $x$, a Content Extractor processes the full image and outputs a content CLS token $\mathbf{c}$ and content patch tokens $\mathbf{T}_c \in \mathbb{R}^{H\times W \times d}$. The global token $\mathbf{c}$ encodes scene semantics and is used to calibrate content patch tokens via a lightweight FiLM modulation, yielding the CSP map:
\begin{equation}
\boldsymbol{\gamma}, \boldsymbol{\beta} = \mathrm{MLP}_c(\mathbf{c}), \quad
\hat{\mathbf{T}}_c = \mathrm{LN}(\boldsymbol{\gamma}\odot \mathbf{T}_c + \boldsymbol{\beta}).
\end{equation}
This calibration stabilizes the content-side query representation across diverse scenes and texture statistics.

In parallel, we partition $x$ into a fixed $4\times4$ grid of crops and feed each crop into a Degradation Extractor to obtain 16 crop-wise degradation CLS tokens $\{\mathbf{d}_i\}_{i=1}^{16}$. We spatially rearrange them into a coarse degradation map $\mathbf{E}\in\mathbb{R}^{4\times4\times d}$ and upsample it to match the patch resolution, producing the DSP map $\mathbf{D}\in\mathbb{R}^{H\times W\times d}$. Importantly, $\mathbf{E}$ captures the \emph{relative} regional distribution of degradation evidence; upsampling resamples this coarse distribution onto the patch grid to serve as a spatial prior for restoration rather than a pixel-accurate degradation estimation.

\paragraph{Content-Guided Degradation Fusion (CGDF).}
With the CSP/DSP maps constructed, we fuse calibrated content patches with the degradation prior tokens using cross-attention:
\begin{equation}
\mathbf{F}' = \hat{\mathbf{T}}_c + \mathrm{Attn}(Q=\hat{\mathbf{T}}_c,\ K=\mathbf{D},\ V=\mathbf{D}),
\end{equation}
followed by a standard LN+FFN block. The resulting $\mathbf{F}'$ forms a \emph{restoration prior map}, providing patch-level guidance signals for subsequent restoration.
Finally, we inject this prior into the restoration backbone via a guided cross-attention mechanism. At layer $l$, we first align the prior $\mathbf{F}'$ to the dimension of the intermediate restoration features $\mathbf{F}_l$, and then use the aligned prior as queries with $\mathbf{F}_l$ as keys/values to produce a guided update, which is merged back to $\mathbf{F}_l$ through a lightweight output projection and a residual connection. This injection enables region-adaptive restoration by letting the prior selectively aggregate backbone responses under spatially varying degradations.

\begin{figure*}[t]
    \centering
    \includegraphics[width=\linewidth,keepaspectratio]{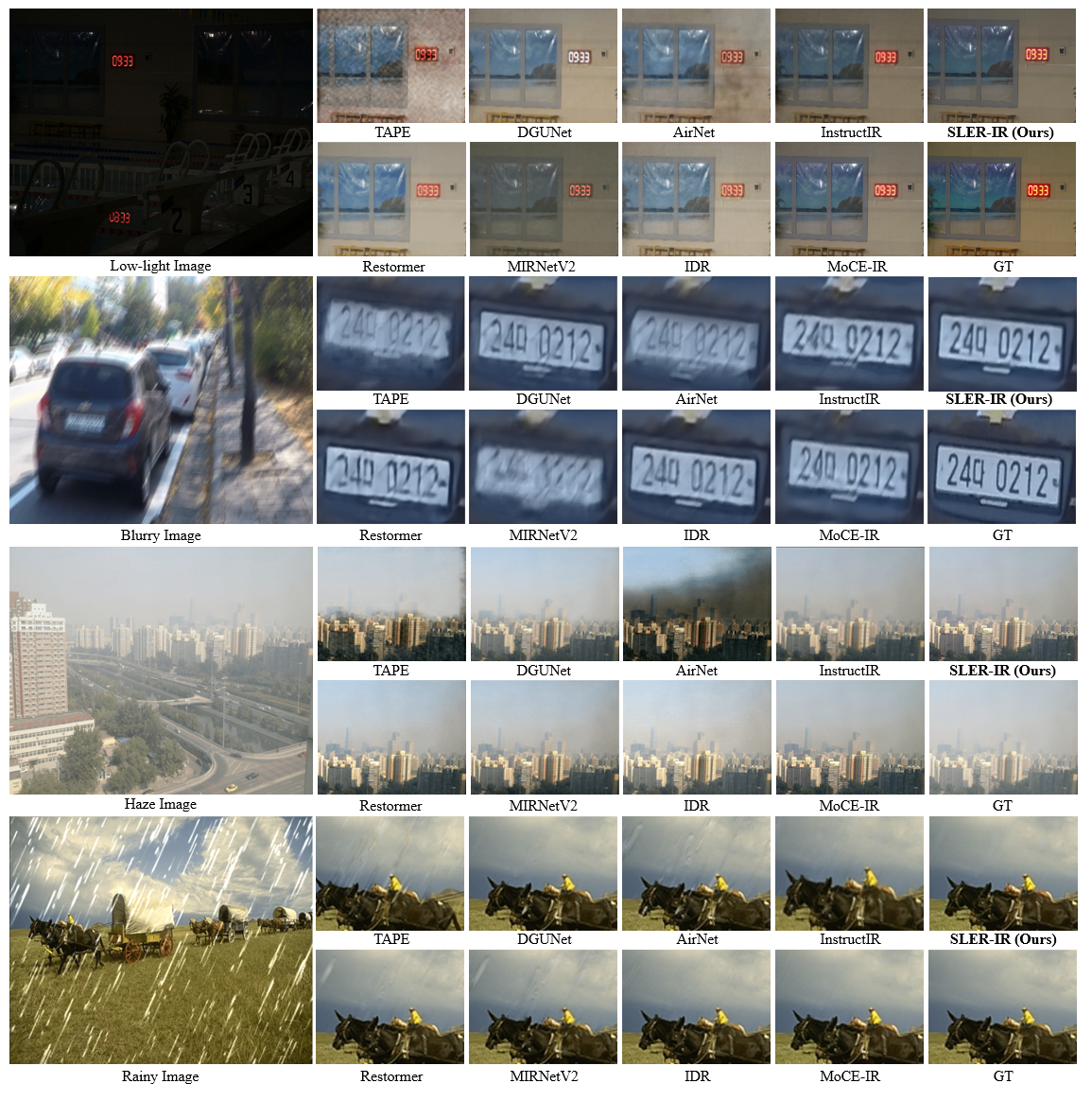}
    \caption{Visual comparison of \textbf{\texorpdfstring{\gradSLERIR}{SLER-IR}} with state-of-the-art methods. Zoom in for better view.
    }
    \label{fig:visual_compare_1}
\end{figure*}

\begin{table*}[t]
\centering
\caption{Quantitative comparison of three-task all-in-one image restoration on SOTS~\cite{li2018benchmarking} (dehazing), Rain100L~\cite{yang2017deep} (deraining), and CBSD68~\cite{martin2001database} (denoising). The best and second-best results are highlighted in \textcolor{red}{red} and \textcolor{blue}{blue}, respectively.}
\label{tab:threetask}

\setlength{\tabcolsep}{6.0pt} % 压列间距（可改 3.0~4.0）
\small % 或 \footnotesize

\begin{adjustbox}{max width=\textwidth}
\begin{tabular}{c c| c c c c c | c}
\toprule
\textbf{Method} & \textbf{Reference} &
\makecell[c]{\textbf{Dehazing}\\\textbf{on SOTS}} &
\makecell[c]{\textbf{Deraining}\\\textbf{on Rain100L}} &
\multicolumn{3}{c|}{\makecell[c]{\textbf{Denoising}\\\textbf{on CBSD68}}} &
\makecell[c]{\textbf{Average}\\\textbf{(PSNR/SSIM)}} \\
\cmidrule(lr){5-7}
& &
\makecell[c]{(PSNR/SSIM)} &
\makecell[c]{(PSNR/SSIM)} &
$\sigma\!=\!15$ & $\sigma\!=\!25$ & $\sigma\!=\!50$ & \\
\midrule
MPRNet & CVPR'21 & 28.00 / 0.958 & 33.86 / 0.958 & 33.27 / 0.920 & 30.76 / 0.871 & 27.29 / 0.761 & 30.63 / 0.894 \\
Restormer & CVPR'22 & 27.78 / 0.958 & 33.78 / 0.958 & 33.72 / 0.865 & 30.67 / 0.865 & 27.63 / 0.792 & 30.75 / 0.901 \\
Airnet & CVPR'22 & 27.94 / 0.962 & 34.90 / 0.967 & 33.92 / 0.933 & 31.26 / 0.888 & 28.00 / 0.797 & 31.20 / 0.910 \\
IDR & CVPR'23 & 29.87 / 0.970 & 36.03 / 0.971 & 33.89 / 0.931 & 31.32 / 0.884 & 28.04 / 0.798 & 31.83 / 0.911 \\
PromptIR & NeurIPS'23 & 30.58 / 0.974 & 36.37 / 0.972 & 33.98 / 0.933 & 31.31 / 0.888 & 28.06 / 0.799 & 32.06 / 0.913 \\

InstructIR & ECCV'24 &
30.22 / 0.959 &
37.98 / 0.978 &
\textcolor{blue}{34.15} / 0.933 &
\textcolor{blue}{31.52} / \textcolor{blue}{0.890} &
\textcolor{blue}{28.30} / \textcolor{blue}{0.804} &
32.43 / 0.913 \\

VLU-Net & CVPR'25 &
30.71 / \textcolor{blue}{0.980} &
\textcolor{red}{38.93} / \textcolor{red}{0.984} &
34.13 / \textcolor{blue}{0.935} &
31.48 / \textcolor{blue}{0.892} &
28.23 / \textcolor{blue}{0.804} &
32.70 / \textcolor{blue}{0.919} \\

MoCEIR & CVPR'25 &
\textcolor{blue}{31.34} / 0.979 &
\textcolor{blue}{38.57} / \textcolor{red}{0.984} &
34.11 / 0.932 &
31.45 / 0.888 &
28.18 / 0.800 &
\textcolor{blue}{32.73} / 0.917 \\

\midrule
\textbf{\texorpdfstring{\gradSLERIR}{SLER-IR}} & Ours &
\textcolor{red}{33.00} / \textcolor{red}{0.982} &
38.34 / \textcolor{blue}{0.983} &
\textcolor{red}{34.25} / \textcolor{red}{0.936} &
\textcolor{red}{31.68} / \textcolor{red}{0.897} &
\textcolor{red}{28.45} / \textcolor{red}{0.813} &
\textcolor{red}{33.14} / \textcolor{red}{0.922} \\
\bottomrule
\end{tabular}
\end{adjustbox}

\end{table*}

\section{Experiment}

This section presents the overall experimental analysis, including the experimental setup, quantitative comparisons with state-of-the-art methods, ablation studies, and qualitative visualizations to comprehensively validate the effectiveness of the proposed framework.

\subsection{Experimental Setup}

\noindent \textbf{Datasets and Evaluation.} Following previous work \cite{li2022all,potlapalli2023promptir}, we prepare datasets for five image degradation tasks, including denoising (BSD400 \cite{arbelaez2010contour} and WED \cite{ma2016waterloo}), dehazing (SOTS \cite{li2018benchmarking}), deraining (Rain100L \cite{yang2017deep}), deblurring (GoPro \cite{nah2017deep}), and low-light enhancement (LOL \cite{wei2018deep}). Following the experimental protocols established in prior works \cite{li2022all, potlapalli2023promptir}, we conduct evaluations under two distinct configurations: (a) the three-task all-in-one and (b) five-task all-in-one setting. For fair comparison, we adopt standard image quality metrics PSNR and SSIM \cite{wang2004image} to evaluate restoration performance. Please refer to supplementary material additional results, visual comparisons, and further analyses.

\noindent \textbf{Implementation Details.} Following the configuration of PromptIR \cite{potlapalli2023promptir}, our \textbf{\texorpdfstring{\gradSLERIR}{SLER-IR}} architecture features a 4-level encoder-decoder structure, comprising varying numbers of Transformer blocks at each level, specifically [4, 6, 6, 8] from level-1 to level-4. We conduct our experiments using PyTorch on H20 GPUs. 
The training is divided into two stages: for Stage 1, we train the model for 15 epochs with a batch size of 10, adopting uniform data sampling. We perform soft branch selection and mask the gradients of non-maximum probability branches. A combined loss function of L1 loss and contrastive loss is utilized to fully enable the model to autonomously learn branch selection for different degradations. The Adam optimizer is employed (\(\beta_1 = 0.9\), \(\beta_2 = 0.999\)) with an initial learning rate of \(2 \times 10^{-4}\), and a cosine decay schedule is adopted. In Stage 2, we adjust the batch size to 20 and train the model for 80 epochs, adopting hard branch selection and using only L1 loss. Throughout the training process, data augmentation operations are employed. 

% Then the model is fine-tuned for 5 epochs at a learning rate of \(2 \times 10^{-5}\) with the patch size adjusted to \(192^2\) and a batch size of 10, followed by 10 epochs at the same learning rate, where the patch size is adjusted to \(224^2\) and the batch size is set to 6.

\subsection{Qualitative and Quantitative Evaluation.}

\noindent \textbf{Three-tasks.} As presented in Table~\ref{tab:threetask}, our method consistently outperforms both general restoration models (MPRNet~\cite{zamir2021multi} and Restormer~\cite{zamir2022restormer}) and recent all-in-one approaches such as AirNet~\cite{li2022all}, PromptIR~\cite{potlapalli2023promptir}, and MoCE-IR~\cite{zamfir2025complexity}.
Specifically, \textbf{\texorpdfstring{\gradSLERIR}{SLER-IR}} achieves the highest average performance of 33.14/0.922 in PSNR/SSIM \cite{wang2004image}, performs better than MoCE-IR by 0.41~dB and 0.005, respectively. 
On the SOTS test set for dehazing, our approach improves PSNR by 1.66~dB over the baseline, and for denoising (CBSD68) it achieves consistent gains across all noise levels ($\sigma$ = 15, 25, 50).

\begin{table*}[t]
\centering
\caption{Quantitative comparison on the five-task all-in-one image restoration setting, including dehazing (SOTS \cite{li2018benchmarking}), deraining (Rain100L \cite{yang2017deep}), denoising (CBSD68 \cite{martin2001database}), deblurring (GoPro \cite{nah2017deep}), and low-light enhancement (LOL \cite{wei2018deep}). The best and second-best results are highlighted in \textcolor{red}{red} and \textcolor{blue}{blue}, respectively.}
\label{tab:fivetask}
\resizebox{\textwidth}{!}{%
\begin{tabular}{c c| c c c c c |c}
\toprule
\textbf{Method} & \textbf{Reference} & 
\makecell[c]{\textbf{Dehazing}\\\textbf{on SOTS}} & 
\makecell[c]{\textbf{Deraining}\\\textbf{on Rain100L}} & 
\makecell[c]{\textbf{Denoising}\\\textbf{on CBSD68}} & 
\makecell[c]{\textbf{Deblurring}\\\textbf{on Gopro}} & 
\makecell[c]{\textbf{Low-light Enh.}\\\textbf{on LOL}} & 
\textbf{Average} \\
\midrule
% NAFNet 数值与第一张表格完全对齐
NAFNet & ECCV’22 & 25.23 / 0.939 & 35.56 / 0.967 & 31.02 / 0.883 & 26.32 / 0.808 & 20.49 / 0.809 & 27.60 / 0.881 \\
FSNet & TPAMI’23 & 25.53 / 0.943 & 36.77 / 0.968 & 31.33 / 0.883 & 28.32 / 0.869 & 22.29 / 0.829 & 28.71 / 0.898 \\
Restormer & ECCV’22 & 24.09 / 0.927 & 34.81 / 0.960 & 31.49 / 0.884 & 27.22 / 0.829 & 20.41 / 0.806 & 27.60 / 0.881 \\
Transweather & CVPR’22 & 21.32 / 0.885 & 29.43 / 0.905 & 29.00 / 0.841 & 25.12 / 0.757 & 21.21 / 0.792 & 25.22 / 0.836 \\
AirNet & CVPR’22 & 21.04 / 0.884 & 32.98 / 0.951 & 30.91 / 0.882 & 24.35 / 0.781 & 18.18 / 0.735 & 25.49 / 0.846 \\

IDR & CVPR’23 &
25.24 / 0.943 &
35.63 / 0.965 &
\textcolor{red}{31.60} / 0.887 &
27.87 / 0.846 &
21.34 / 0.826 &
28.34 / 0.893 \\

PromptIR & NeurIPS’23 & 26.54 / 0.949 & 36.37 / 0.970 & 31.47 / 0.886 & 28.71 / 0.881 & 22.68 / 0.832 & 29.15 / 0.904 \\

InstructIR & ECCV’24 &
26.14 / 0.973 &
37.10 / 0.966 &
31.44 / 0.887 &
29.40 / 0.886 &
\textcolor{blue}{23.00} / 0.836 &
29.55 / 0.907 \\

VLU-Net & CVPR’25 &
\textcolor{blue}{30.84} / \textcolor{blue}{0.980} &
\textcolor{red}{38.54} / \textcolor{blue}{0.982} &
31.43 / \textcolor{blue}{0.891} &
27.46 / 0.840 &
22.29 / 0.833 &
30.11 / 0.905 \\

MoCEIR & CVPR’25 &
30.48 / 0.974 &
38.04 / \textcolor{blue}{0.982} &
31.34 / 0.887 &
\textcolor{blue}{30.05} / \textcolor{blue}{0.899} &
\textcolor{blue}{23.00} / \textcolor{blue}{0.852} &
\textcolor{blue}{30.58} / \textcolor{blue}{0.919} \\

\midrule
\textbf{\texorpdfstring{\gradSLERIR}{SLER-IR}} & Ours &
\textcolor{red}{33.43} / \textcolor{red}{0.985} &
\textcolor{blue}{38.47} / \textcolor{red}{0.983} &
\textcolor{blue}{31.50} / \textcolor{red}{0.894} &
\textcolor{red}{31.27} / \textcolor{red}{0.918} &
\textcolor{red}{23.96} / \textcolor{red}{0.861} &
\textcolor{red}{31.73} / \textcolor{red}{0.928} \\
\bottomrule
\end{tabular}%
}
\end{table*}

\noindent \textbf{Five-tasks.} As shown in Table~\ref{tab:fivetask}, \textbf{\texorpdfstring{\gradSLERIR}{SLER-IR}} still achieves the best performance, which attains the highest overall average of 31.73/0.928 in PSNR/SSIM \cite{wang2004image}, surpassing the previous approach by a margin of 1.15~dB and 0.009, respectively. 
Notably, the proposed method achieves remarkable gains of +2.59~dB on the SOTS dataset for dehazing and +1.22~dB on GoPro for deblurring. Even on the challenging low-light enhancement task (LOL), our model maintains a clear advantage with an improvement of +0.96~dB.
These results demonstrate the effectiveness and robustness of the proposed routing framework across diverse restoration tasks.

\noindent \textbf{Quantitative Results.} 
% Figure \ref{fig:visual_compare} presents visual results across dehazing, denoising, and deraining. In hazy scenes, AirNet \cite{li2022all} and PromptIR \cite{potlapalli2023promptir} leave residual haze with color deviations, while our method achieves accurate color and haze removal. For rainy scenes, competing methods retain rain streaks, which our approach fully eliminates. In denoising, our method produces sharp, clear outputs with no color deviation. These results, coupled with quantitative comparisons, demonstrate the effectiveness of our method.
Figure~\ref{fig:visual_compare_1} shows visual comparisons on low-light enhancement, deblurring, dehazing, and deraining tasks. In low-light scenes, competing methods often produce insufficient brightness or lose fine structural details, whereas SLER-IR achieves balanced enhancement with well-preserved structures and natural illumination. For deblurring, existing approaches often leave noticeable residual blur, while our method restores clearer textures and sharper edges. In dehazing, other methods tend to retain haze artifacts and color distortions, whereas \textbf{\texorpdfstring{\gradSLERIR}{SLER-IR}} delivers cleaner results with more faithful color reconstruction. Under rainy conditions, the proposed model removes rain streaks more thoroughly while preserving underlying scene details. These visual results, together with the quantitative comparisons, further validate the effectiveness and robustness of our approach.

% ----- Row 1: loss + model -----
\begin{table}[t]
\centering
\footnotesize
\setlength{\tabcolsep}{5pt}

\begin{minipage}[t]{0.49\linewidth}
\centering
\caption{Ablation study on loss functions.}
\label{tab:ablation_loss}
\begin{tabular}{l p{1.7cm} S[table-format=2.2] S[table-format=1.3]}
\toprule
Loss & Branch Sel. & {PSNR} & {SSIM} \\
\midrule
$L_1$ & Classifier & 28.30 & 0.879 \\
$L_1+L_{\text{cls}}$ & Classifier & 28.76 & 0.884 \\
$L_1+L_{\text{HC}}$ & Ours & \bfseries 31.73 & \bfseries 0.928 \\
\bottomrule
\end{tabular}
\end{minipage}\hfill
\begin{minipage}[t]{0.49\linewidth}
\centering
\caption{Ablation study on different model components.}
\label{tab:ablation_model}
\begin{tabular}{p{2.6cm} S[table-format=2.2] S[table-format=1.3]}
\toprule
Model & {PSNR} & {SSIM} \\
\midrule
w/o GLGF & 30.64 & 0.917 \\
SLER-IR & \bfseries 31.73 & \bfseries 0.928 \\
\bottomrule
\end{tabular}
\end{minipage}
\end{table}

\vspace{-2mm}

% ----- Row 2: expert num + alpha -----
\begin{table}[t]
\centering
\footnotesize
\setlength{\tabcolsep}{7pt}

\begin{minipage}[t]{0.49\linewidth}
\centering
\caption{Ablation study on the number of experts per layer.}
\label{tab:ablation_expert_num}
\setlength{\tabcolsep}{19pt}
\begin{tabular}{c S[table-format=2.2] S[table-format=1.3]}
\toprule
$C$ & {PSNR} & {SSIM} \\
\midrule
2 & 31.22 & 0.924 \\
3 & \bfseries 31.73 & \bfseries 0.928 \\
4 & 31.32 & 0.926 \\
\bottomrule
\end{tabular}
\end{minipage}\hfill
\begin{minipage}[t]{0.49\linewidth}
\centering
\caption{Ablation study on hyperspherical contrastive loss weight ($\alpha$).}
\label{tab:ablation_cont_weight}
\setlength{\tabcolsep}{19pt}
\begin{tabular}{c S[table-format=2.2] S[table-format=1.3]}
\toprule
$\alpha$ & {PSNR} & {SSIM} \\
\midrule
0.5 & 31.03 & 0.923 \\
1 & \bfseries 31.73 & \bfseries 0.928 \\
2 & 31.15 & 0.924 \\
\bottomrule
\end{tabular}
\end{minipage}
\end{table}

\subsection{Ablation Studies}

To validate the effectiveness of the proposed components, we conduct ablation experiments under the five-task setting. Additional results and analyses are provided in the supplementary material.

\noindent \textbf{Effectiveness of the proposed losses.} Table~\ref{tab:ablation_loss} evaluates the impact of different training objectives. Starting from the baseline model trained with the restoration loss ($L_1$), introducing the auxiliary classifier loss improves performance by providing additional degradation-aware supervision. Further incorporating the proposed hyperspherical contrastive loss yields the best PSNR and SSIM results. This demonstrates that contrastive degradation representation learning helps enhance the discriminability of degradation embeddings, leading to more reliable expert routing and improved restoration performance.

\noindent \textbf{Impact of Global–Local Granularity Fusion (GLGF).} Table~\ref{tab:ablation_model} evaluates the effectiveness of the proposed GLGF module. As shown in the results, integrating GLGF consistently improves restoration performance compared with the baseline model without this module. This improvement indicates that jointly modeling global semantic cues (CLS tokens) and local degradation evidence (patch-level cues) provides more informative restoration guidance via the learned restoration prior.

\noindent \textbf{Ablation of Expert Number per Layer.} Table~\ref{tab:ablation_expert_num} investigates the influence of the number of experts per layer ($C\in\{2,3,4\}$). With identical training epochs and experimental settings, $C{=}3$ achieves the best trade-off between expert specialization and computational efficiency. Fewer experts limit discrimination capacity, while more experts increase computation without clear gains under the same training budget.

\noindent \textbf{Ablation of Hyperspherical Contrastive Loss Weight.} Table~\ref{tab:ablation_cont_weight} analyzes the influence of the hyperspherical contrastive loss weight $\alpha$. When the weight is too small ($\alpha{=}0.5$), the degradation embedding receives insufficient supervision, resulting in less discriminative routing representations. Conversely, an excessively large weight ($\alpha{=}2$) overemphasizes contrastive optimization and weakens the restoration objective. Empirically, $\alpha{=}1$ achieves the best balance between degradation representation learning and restoration performance, and is therefore used as the default setting.

\label{sec:experiment}

% ----------------- 表格 1 -----------------

% ----------------- 表格 2（修正后） -----------------
% ----------------- 表格 2（最终完全修正版） -----------------

\section{Conclusion}
\label{sec:conclusion}

In this paper, we propose \textbf{\texorpdfstring{\gradSLERIR}{SLER-IR}}, a spherical layer-wise expert routing framework for unified all-in-one image restoration. By leveraging hyperspherical degradation embedding for geometry-consistent routing and a global–local granularity fusion module for spatially robust guidance, our model enables progressive expert specialization under diverse degradations and improves adaptability to complex degradation scenarios. In addition, the proposed routing mechanism allows flexible compositional inference paths without increasing inference overhead. Extensive experiments on multiple benchmarks demonstrate that our approach consistently outperforms state-of-the-art methods in both quantitative metrics and visual quality.

% \noindent \textbf{Limitations.} Despite the promising results achieved by \textbf{\texorpdfstring{\gradSLERIR}{SLER-IR}}, several limitations remain and provide directions for future research. Our routing mechanism relies on the quality of the learned degradation embeddings. Although hyperspherical embedding improves routing stability, the performance may still depend on how accurately degradation characteristics are captured. Exploring more expressive degradation representations or self-supervised degradation discovery could further enhance routing robustness. In addition, while SLER-IR demonstrates strong performance across multiple degradation types, the current framework is evaluated on a limited set of restoration tasks. Extending the routing mechanism to a broader range of degradations or more complex restoration scenarios, remains an interesting direction for future work.

\bibliographystyle{splncs04}
\bibliography{main}
\end{document}